# Enhanced Classification Accuracy for Cardiotocogram Data with Ensemble Feature Selection and Classifier Ensemble

**Tipawan Silwattananusarn[1], Wanida Kanarkard[2], Kulthida Tuamsuk[3]**

[1]Department of Library and Information Science, Faculty of Humanities and Social Sciences, Prince of Songkla University, Hat Yai, Thailand
[2]Department of Computer Engineering, Faculty of Engineering, Khon Kaen University, Khon Kaen, Thailand
[3]Information and Communication Management Program, Faculty of Humanities and Social Sciences, Khon Kaen University, Khon Kaen, Thailand
Email: tipawan.s@psu.ac.th, wanida@kku.ac.th, kultua@kku.ac.th

## Abstract

**In this paper ensemble learning based feature selection and classifier ensemble model is proposed to improve classification accuracy. The hypothesis is that good feature sets contain features that are highly correlated with the class from ensemble feature selection to SVM ensembles which can be achieved on the performance of classification accuracy. The proposed approach consists of two phases: (i) to select feature sets that are likely to be the support vectors by applying ensemble based feature selection methods; and (ii) to construct an SVM ensemble using the selected features. The proposed approach was evaluated by experiments on Cardiotocography dataset. Four feature selection techniques were used: (i) Correlation-based, (ii) Consistency-based, (iii) *ReliefF* and (iv) Information Gain. Experimental results showed that using the ensemble of Information Gain feature selection and Correlation-based feature selection with SVM ensembles achieved higher classification accuracy than both single SVM classifier and ensemble feature selection with SVM classifier.**

## 1. Introduction

In recent years, research and development have been going on in healthcare industry and will grow in the collection and analysis of all pertinent data in healthcare e.g. biotechnology, medicine and biomedical. Thus the research in the healthcare area must ensure safety and reliability. The researcher or scientists must make sure that the model developed has the highest level of accuracy of predicting the outcomes. As these models are involved in predicting the highly precise attributes related to human life, a perfect diagnosis has to be provided based on the outcome.

The major challenge in medical domain is the extraction of comprehensible knowledge from medical diagnosis data such as CTG data. Cardiotocography (CTG) is used to evaluate fetal well-being during the delivery. Many researchers have employed different machine learning methods to help the doctors to interpret the CTG trace pattern and to act as a decision support system in obstetrics. The effectiveness of classification and recognition systems has improved in a great deal to help medical experts in diagnosing diseases. **Table 1** illustrates the current works in CTG data that used the data mining functions of classifier methods. From the review articles published on this subject, it was found that there is no identity on the best methodology for baseline estimation in prediction and classification on CTG data. So, in this work, we are going to evaluate some of data mining techniques for the classification of CTG data.

Most people are often prone to making mistakes during analyses or, possibly, when trying to establish relationships between multiple features. This makes it difficult for them to find solutions to certain problems. Machine learning can often be successfully applied to these problems, improving the efficiency of systems and the designs of machines. The large data sets must be analyzed and interpreted to extract all the relevant information they can provide. Feature selection and classification techniques are the main tools to pursue this task. Feature selection techniques are meant to identify a small subset of important data within a large data set. Classification techniques are designed to identify and synthetic models are able to explain some of the relevant characteristics contained. The two techniques are indeed strongly related: the selection of a few relevant features among the many and the application of learning. Feature selection is always used on large bodies of data to identify those features on which to apply a classification method to identify meaningful models.

Classification is a core task in data mining and is the process of finding models, based on selected features from training data that divide a data item into one of a number of classes [6] [7]. In recent years, researchers have been using the ensemble classifiers in machine learning to improve classification accuracy and reduce the time consumed.

Ensemble classification has received much attention in machine learning and has demonstrated capabilities in improving classification accuracy. The support vector machines ensemble has been applied in many areas to improve classification accuracy performance. There are at least three reasons for the success of SVMs: their ability to learn well with only a very small number of free parameters; their robustness against several types of model violations and outliers, and their computational efficiency compared with several other methods [8]. For example, an ensemble of SVM classifiers has been applied for reliable classification of biological data [9]; an empirical research into SVM ensembles [10]; and a novel algorithm for constructing a support vector machines classification ensemble [11].

The idea of an ensemble in biological data originates from combining multiple classifiers for improving sample classification accuracy. As a consequence of the growth in the instability of the feature selection results from high-dimensional data, it has recently been adapted and increasingly used in feature selection. For example, an ensemble of feature selection techniques for high dimensional data [12] has been used for improving classification accuracy by using feature selection and an ensemble model [13]; a novel ensemble machine learning for robust microarray data classification [14]; feature selection for cancer classification with an SVM based approach [15]; and on the relationship between feature selection and classification accuracy [16].

**Table 1.** Representative works that used the functions of feature selection methods and classifier methods.

| Reference | FS Methods | Classifier Methods |
|---|---|---|
| [1] | N/A | Adaptive Boosting, Decision Tree |
| [2] | N/A | Discriminant Analysis, Decision Tree, Artificial Neural Network |
| [3] | N/A | SVM, ANN, Fuzzy c-mean, Clustering |
| [4] | N/A | Artificial Neural Network |
| [5] | N/A | Radial Basis Function Networks, Ensemble Learning |

The Support Vector Machine is one machine learning algorithm which can solve classification problems, use a flexible representation of the class boundaries, implement automatic complexity control to reduce over-fitting, and has a single global minimum which can be found in polynomial time. It is popular because it can be easy to use, it often has good generalization performance and the same algorithm solves a variety of problems with little tuning [12] [16]-[18].

The SVM has been successfully applied in many areas such as pattern recognition, face detection, bioinformatics; however, the SVM has two disadvantages. First, using a combination of SVMs for the multi-class classification does not seem to improve the performance as much as in the binary classification. Second, the SVM learning is very time consuming for a large scale of data [19]. An experiment in this study used DNA sequences data which is characterized by small sample size, high dimensionality, noise and large instability.

In this study, we decide to step into this challenging arena to come up with a proposed method on Cardiotocography. Cardiotocography is a recording of the fetal heartbeat and the uterine contractions during the pregnancy period. Physicians and specialists spend large amounts of time on each and every Cardiotocography report analysis. To improve the capability on these report analyses and to provide a proper diagnosis to a patient with those symptoms, we have developed a proposed model that would take 23 input variables from the Cardiotocography data set to predict the classification of fetal heart rate which is an output variable. This model classifies the output variable into three classes: N-Normal, S-Suspect and P-Pathologic. As the treatment provided to the patients would be based on the classification of fetal heart rate, we define our accuracy level as a very high 95% to make a reliable model.

## 2. Material and Methodology

This study focuses on the determinant approaches of ensemble feature selection and a classifier ensemble to CTG data classification. It examines the ensemble learning on two feature selection techniques: feature subset selection and feature rankers. We evaluate four different strategies for feature selection: (i) Correlation-based feature selection, (ii) Consistency-based feature selection, (iii) *ReliefF*, and (iv) Information Gain. A bagging ensemble is considered in order to evaluate our feature selection technique. Support Vector Machine is defined as our basic classifier learning. In this study, the bagging method for constructing the component SVM and a majority voting strategy for aggregating several trained SVMs are applied. An overview of the research approaches is shown IN **Figure 1**.

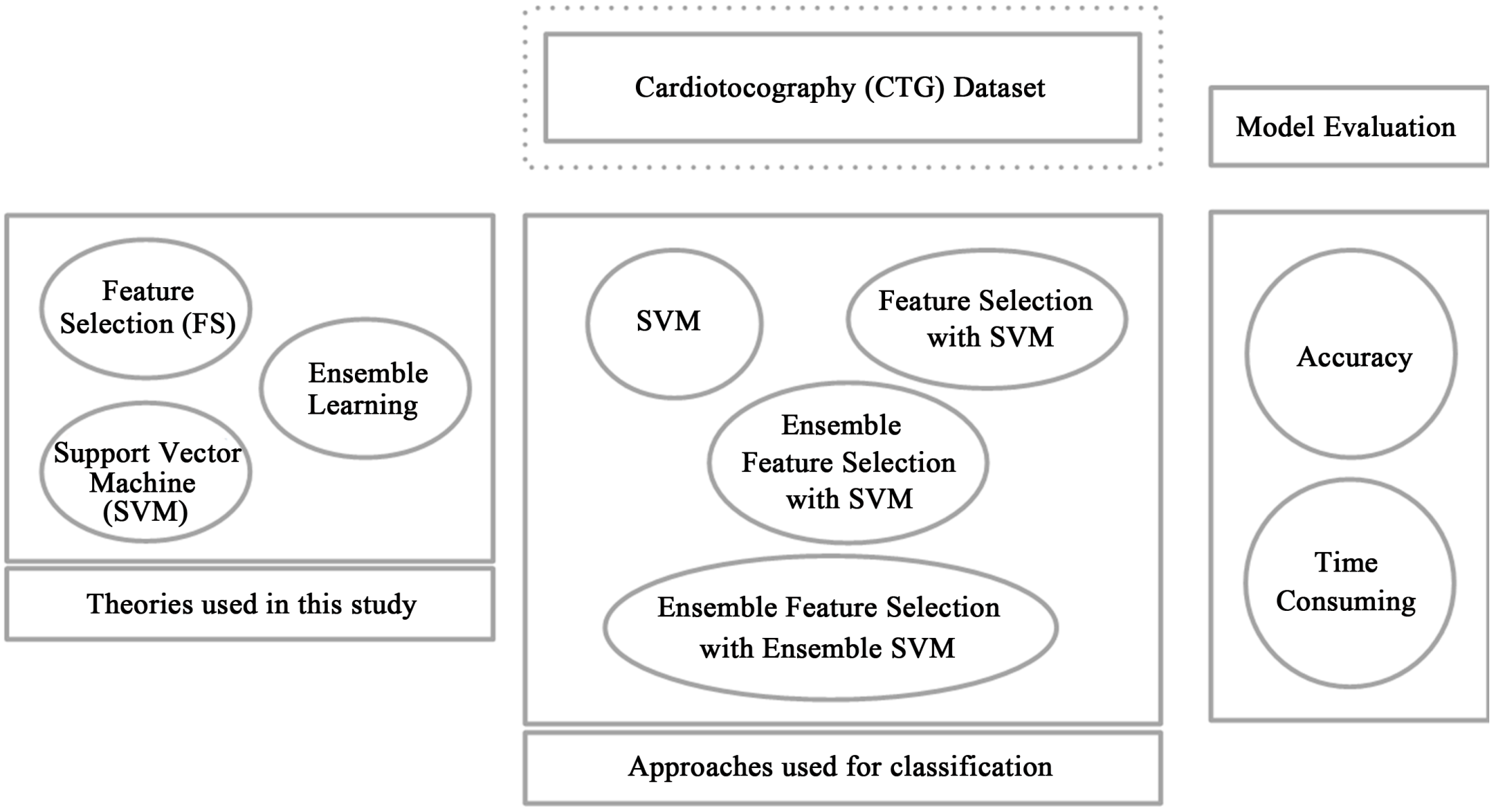

**Figure 1.** Overview of research approaches.

### 2.1. Dataset Descriptions

Experiments conducted in this study were performed on a Cardiotocography (CTG) dataset which was collected from the UCI Machine Learning Repository, which is available at http://archive.ic.uci.edu/ml/. The Cardiotocography dataset consisted of 23 attributes and 2126 instances. All attributes were numeric. A class attribute for the Cardiotocography dataset had 3 distinct values: *Normal, Suspect, and Pathologic*.

***The Cardiotocography*** (***CTG***) ***dataset*** consisted of the measurement of Fetal Heart Rate (FHR) and Uterine Contraction (UC) features on Cardiotocograms data, used to evaluate maternal and fetal well-being during pregnancy and before delivery [4]. We used this dataset for these evaluations.

*Attribute Information*:

1. LB–FHR baseline (beats per minute)
2. AC-# of accelerations per second
3. FM-# of fetal movements per second
4. UC-# of uterine contractions per second
5. DL-# of light decelerations per second
6. DS-# of severe decelerations per second
7. DP-# of prolonged decelerations per second
8. ASTV–percentage of time with abnormal short term variability
9. MSTV–mean value of short term variability
10. ALTV–percentage of time with abnormal long term variability
11. MLTV–mean value of long term variability
12. Width–width of FHR histogram
13. Min–minimum of FHR histogram
14. Max–Maximum of FHR histogram
15. Nmax-# of histogram peaks
16. Nzeros-# of histogram zeros
17. Mode–histogram mode
18. Mean–histogram mean
19. Median–histogram median
20. Variance–histogram variance
21. Tendency–histogram tendency
22. CLASS–FHR pattern class code (1 to 10)
23. NSP–fetal state class code (N = Normal; S = Suspect; P = Pathologic)

*Class Information*: The descriptions of a three-class fetal state classification are:

| Class Information | Description |
|---|---|
| Normal | All four features fall into the reassuring category. |
| Suspicious | Features fall into one of the non-reassuring categories and the reassuring category and the remainder of the features are reassuring. |
| Pathological | Features fall into two or more of the non-reassuring categories and the reassuring category or two or more abnormal categories |

Visualization of the Data Patterns: **Figure 2** roughly shows the projection of 23 attributes data into a two dimensional data space. In this plot, the normal CTG data points are shown in black, the suspicious data points are shown as red, and the pathologic data points are shown as green.

### 2.2. Experimental Design and Methodology

#### 2.2.1. Feature Selection

Feature selection can be divided into feature subset selection and feature ranking. Feature subset selection selects a subset of attributes which collectively increases the performance of the model. Feature ranking calculates the scores of each attribute and then sorts them according to their scores. In this work we focus primarily on two filter based feature subset selection techniques and two filter based feature ranking techniques. They are (i) Correlation-based, (ii) Consistency-based, (iii) *ReliefF*, and (iv) Information Gain.

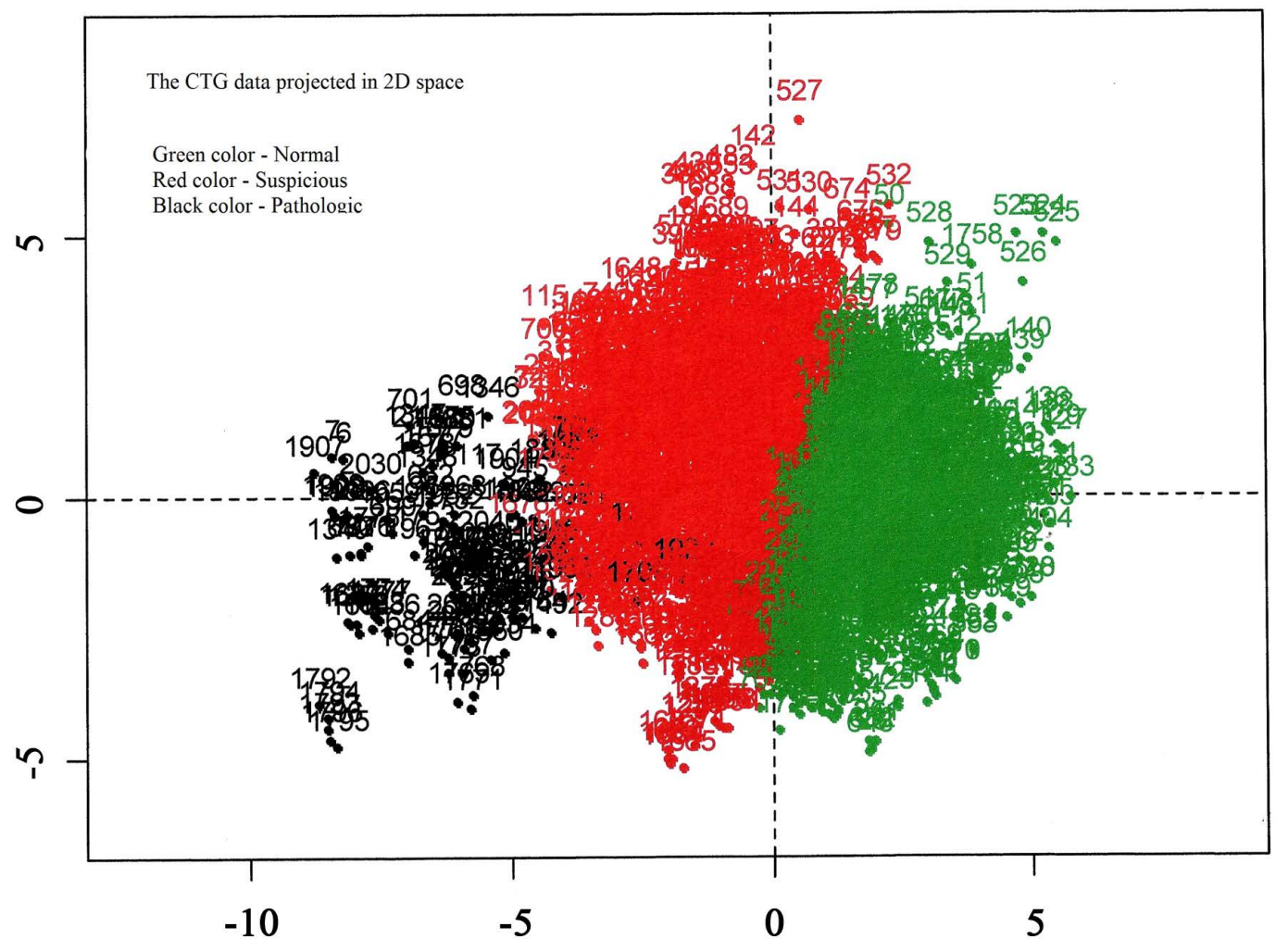

**Figure 2.** The CTG data projection in 2D space.

**1. Feature Subset Selection Techniques**

**1) Correlation-based Feature Selection**

This is a simple filter algorithm that ranks feature subsets according to a correlation-based heuristic evaluation function [20]. The bias of the evaluation function is toward subsets that contain features that are highly correlated with the class and uncorrelated with each other. Irrelevant features should be ignored and redundant features should be screened out.

**2) Consistency-based Filter**

The consistency-based filter [20] evaluates the worth of a subset of features by the level of consistency in the class values when the training instances are projected onto the subset of attributes. The algorithm generates a random subset S from the number of features in every round. If the number of features of S is less than the current best, the data with the features prescribed in S is checked against the inconsistency criterion. If its inconsistency rate is below a pre-specified one, S becomes the new current best.

**1. Feature Ranking Techniques**

**1) *ReliefF***

*ReliefF* [21] is an extension of the original Relief algorithm [22] that adds the ability to deal with multiclass problems and it is more robust and capable of dealing with incomplete and noisy data. The Relief family of methods are attractive because they may be applied in all situations, have low bias, include interaction among features and may capture local dependencies that other methods miss.

**2) Information Gain**

One of the most common attribute evaluation methods is called Information Gain [23]. This method provides an ordered ranking of all the features and then a threshold is required. Information Gain is based on the concept of entropy. The expected value of information gain is the mutual information of the target variable.

#### 2.2.2. Support Vector Machines

A SVM is one of the standard tools for machine learning and data mining and is found in many applications in the healthcare area and bioinformatics [24]. A Support Vector Machine (SVM), which was first developed in 1992 by Vapnik [25], is a learning method that is trained on a whole training dataset. The idea behind the SVM algorithm comprises four basic concepts: (i) the separating hyperplanes, (ii) the maximum-margin hyperplane, (iii) the soft margin, and (iv) the kernel function [26]. SVM performs classification tasks by selecting a small number of critical boundary instances called support vectors from each class and constructing hyperplanes in a multidimensional space that separates cases of different class labels [27]. Next, SVM employs a set of training data into a suitable space and then learns a function to separate the data with a hyperplane [28]. Because SVM is able to minimize the probability of error by using the trained model that generalizes well on unseen data, SVM has been widely applied within the field of computational biology [29].

A support vector machine algorithm works as follows. It tries to find the line, called the maximum-hyperplane, which separates the tuples of one class from another. The dividing line has been chosen so that the parallel lines that touch the items from each class are as far from it as possible. The distance between the parallel lines is called the margin. The points at the margin are needed to determine the placement of the dividing line. The points near the line are called the support vectors [27] [30]. Thus, SVM finds the support vectors and uses them to find the dividing line with maximum distance between the nearest training tuples [30].

Instead of fitting nonlinear curves to the data, SVM handles this by using a kernel function to map the data into a different space where a hyperplane can be used to do the separation. The kernel function may transform the data into a higher dimensional space to perform the separation. It allows SVM models to perform separations even with very complex boundaries [30]. The effectiveness of SVM depends on the kernel selection, the kernel's parameters, and the soft margin parameter [30].

There are two main motivations to widely use SVM with related kernel methods in computational biology. First, SVM involves only a few samples in the determination of the classification function and can deal with high-dimensional, noisy, and large datasets when compared to other statistical or machine learning methods [25] [27]. Second, the most common types of data in biology applications are not vector inputs, for example, variable length sequences or graphs. The SVM can handle the non-vector inputs [25] [26]. Although SVM would be good to consider as many training datasets as possible, the disadvantage of this method is that it is time consuming when supplied with test datasets [31].

**Figure 3** shows the block diagram of the experimental design. The experimental dataset in this study is the Cardiotocography (CTG) dataset. The suite of filters employed contained four filters based on various feature search strategy approaches. These are Correlation-based feature selection (CorrFS), Consistency-based feature selection (ConFS), *ReliefF* (RFF) and Information Gain (IG). Two feature search techniques were applied in this study *i.e.* BestFirst search technique and GeneticSearch technique. The two remaining feature selections of *ReliefF* and Information Gain are based feature ranking techniques. The ensemble approach has been applied to various feature selections.

The experiments compare runs of the experimental model which is machine learning algorithms with and without feature selection on the datasets. In each trial, a model is trained on the training dataset and is evaluated on the test set. When models are compared, each is applied to the same training and test set. The testing accuracy of a model is the percentage of test examples it classifies correctly. A 70% of training and 30% of testing split was used in CTG data.

### 2.2.3. Ensemble Learning Based Feature Selection

An ensemble learning based feature selection approach has been proposed to improve the stability of feature selection [32]. The idea is to generate multiple diverse feature selectors and combine their outputs. This approach has the ability to handle stability issues superior to existing feature selection methods. The stability of feature selection is extremely important in bioinformatics and stable feature selection will remain relevant even when changes to the data occur [33].

There are two essential steps in creating a feature selection ensemble. The first step is to generate a set of diverse feature selectors. Then, the second step is to aggregate the results of these feature selectors. The components in ensemble learning based feature selection are shown in **Figure 4**.

Two of the bigger problems associated with medical diagnosis data are the high dimensionality and low sample size of the resulting datasets. High dimensionality occurs when there are a large number of features per instance of data. These combined problems cause the stability of feature selectors to decrease.

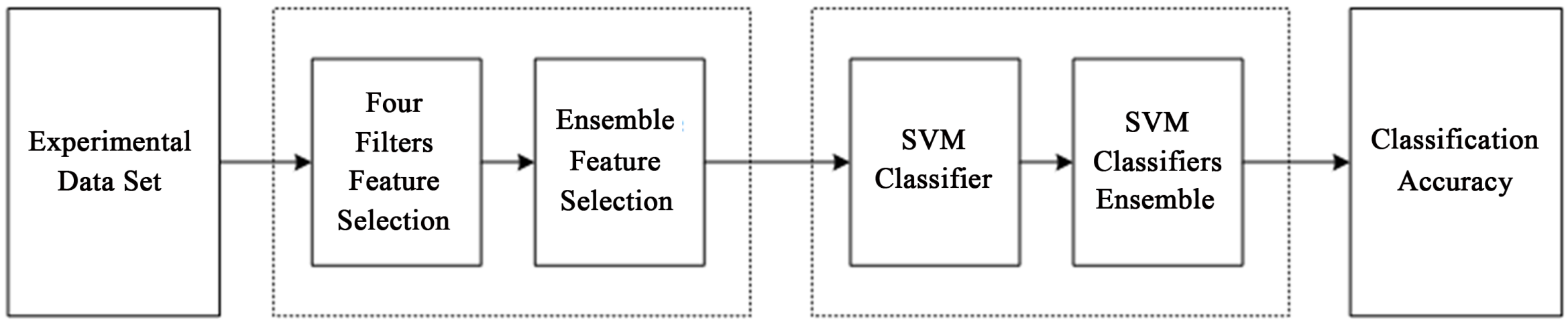

**Figure 3.** Block diagram of the experimental design.

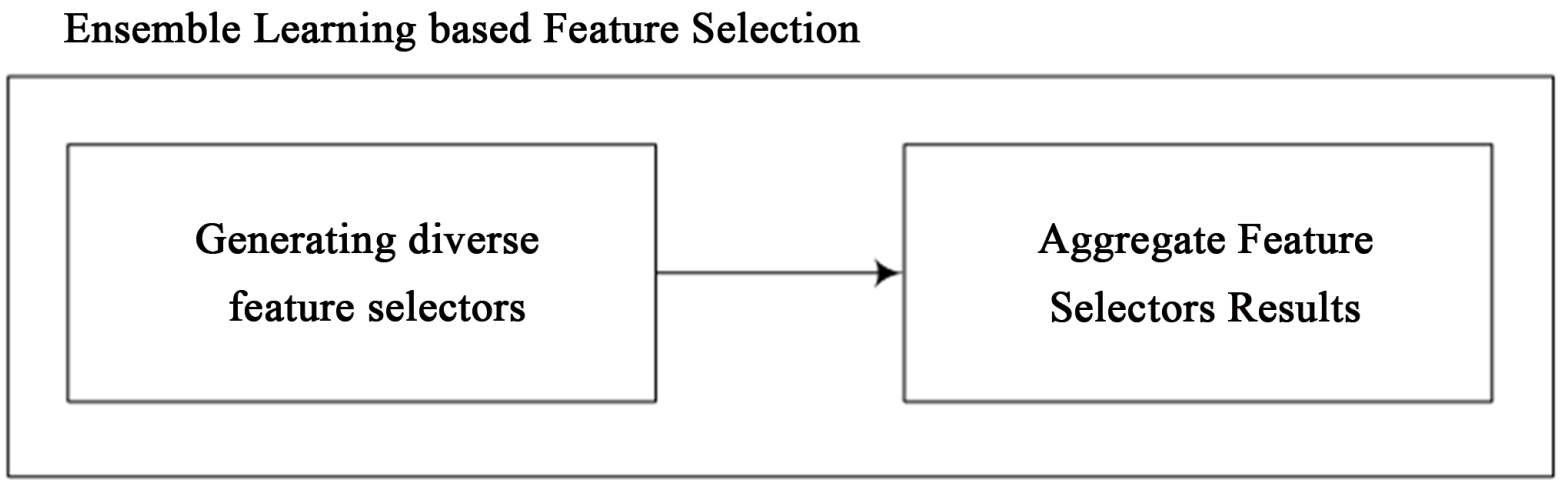

**Figure 4.** Components of ensemble learning based feature selection.

#### 2.2.4. Ensemble Learning

Ensemble methods are learning algorithms that construct a set of base classifiers to combine and then classify new data points by taking a vote on their predictions [6] [27]. The learning procedure for ensemble algorithms is divided into two sections. The first is a construction of the base classifiers, and the second is for a voting task. The vote is used to combine classifiers. There are various kinds of voting systems and the two main types are weighted voting and un-weighted voting. In the weighted voting system, each base classifier holds different voting power. In the un-weighted voting, the individual base classifier has equal weight, and the winner is the one with the most votes [27]. We used an ensemble method to combine classifiers as an ensemble where each classifier is given a weight according to the order of feature sets selected and their training errors. The ensemble method performs classification by giving the weight of the predictions made by each best classifier.

**1) Ada Boost**

Ada Boost, is a machine learning algorithm underlying theory of Boosting, introduced by Freund and Schapire in 1995 [34]. The idea behind the Adaboost algorithm is to maintain repeatedly a set of weights over the training set in a series of iterations and combine into a weighted majority vote in order to achieve a higher accuracy [35] [36]. Initially, the same weight is assigned to all instances. On each round, the weights of each incorrectly classified instance are increased while the weights of correctly classified instance are decreased. A weight is assigned to several individual classifiers. The weight measures the total weight of the correctly classified instances. Thus, higher weights are given to more accurate classifiers. These weights are used for the classification of new instances [35] [36]. Furthermore, the most basic property of AdaBoost concerns reducing the training error [36]. From a number of studies of AdaBoost, it appears to have larger error reductions and tends to reduce both bias and variance in terms of error performance [36].

A general method of AdaBoost is to improve the performance of a weak learner by repeatedly running it on various training data and then combining the weak learners into a single classifier that outperforms every one of them [35]. The advantages of the AdaBoost method are that it is fast, simple, easy to program and there are no parameters to set (except for the number of iterations) [36]. AdaBoost can handle noisy data and identify outliers; however, it can fail to improve the performance of the base classifiers when there are insufficient data [36].

**2) Bagging**

Bagging (stands for **B**ootstrap **Agg**regat**ing**) is a machine learning algorithm underlying the theory of bagging predictors, introduced by Leo Breiman in 1996 [34]. The idea of a Bagging algorithm is to generate bootstrap samples of a training data using sampling with replacement. Each bootstrap sample is used to train a different component of a base classifier. The outputs of the models are combined by majority vote of each component classifier to create a single output. Bagging improves the performance for classifier instability which varies significantly with small changes in the training set [34].

Support Vector Machine has been widely applied to many areas of healthcare and is becoming popular in a wide variety of medical applications. This is because SVM has been demonstrated to have a very powerful and strong generalization performance. It is designed to maximize the margin to separate two classes so that the trained model generalizes well on unseen data [35].

In constructing the SVM ensemble it is most important that each individual SVM becomes as different from another SVM as possible [19]. The idea of the Support Vector Machine ensemble with bagging and boosting has been proposed in [19] to improve the limited classification performance of the single SVM..

**Figure 5** shows the architecture of the experimental model. Feature selection techniques are applied to the dataset to reduce unimportant attributes and then applied to the SVM classifier and the SVM ensembles model. The experimental dataset is partitioned into 70% of training set and 30% of testing set. The performances of the classifiers and ensemble model are measured by classification accuracy.

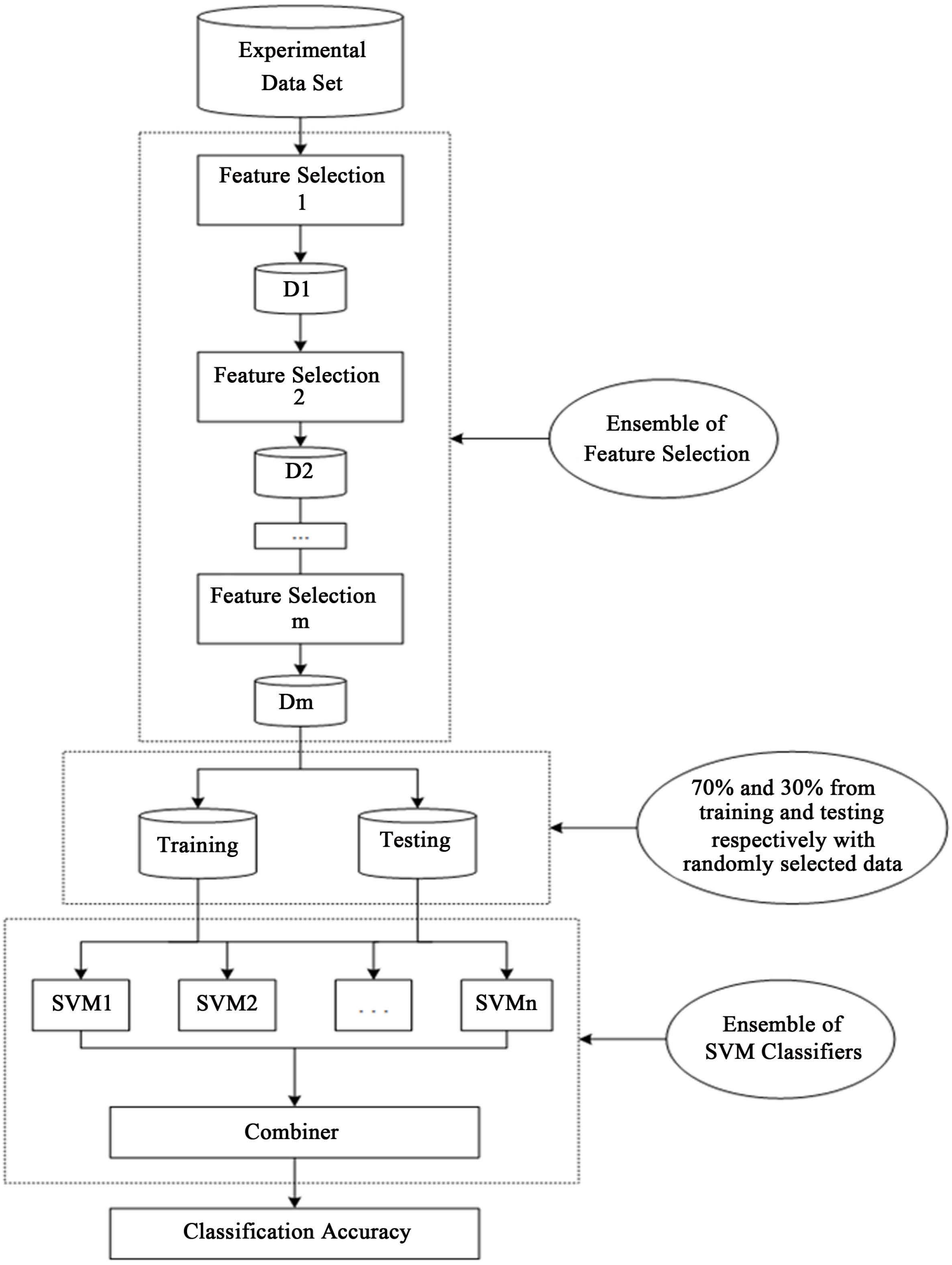

**Figure 5.** Architecture of the experimental model.

The parameters of the feature selection technique used in the experimental framework for the CTG data are illustrated in **Table 2**.

# 3. Experimental Results

## 3.1. Performance Evaluation of SVM on CTG Data

Our first set of experiments was focused on evaluating the performance of the Support Vector Machine algorithm on the Cardiotocography (CTG) data. We divided the data into two sets: the first group was set with 70 percent of the source data, for training the model, and the other group with 30 percent of the source data, for testing the model. **Table 3** shows the computed confusion matrix of the SVM model with polynomial kernel function for C = 10 and degree-3 of training and test data partition before feature selection.

Each cell in **Table 3** contains the row number of samples classified for the corresponding combination of desired output and actual output with splitting 70% for training data and the remainder for the test data. **Table 4** shows the values of statistical parameters (classification accuracy) for the SVM model of training and test data partition before feature selection. **Table 5** shows the output from the SVM classifier model before feature selection.

**Table 2.** Parameters of the feature selection techniques.

| Feature Selection | Feature Selection Techniques Used | |
|---|---|---|
| | Searching Strategy | Evaluation Criterion |
| Correlation-based | Filter | Genetic Search |
| Consistency-based | Filter | Genetic Search |
| *ReliefF* | Filter | Ranker |
| Information Gain | Filter | Ranker |

**Table 3.** Confusion matrix of the SVM (C = 10, degree-3) model of training and test data partition before feature selection.

| Model | Desired Output | Actual Output | | | | | |
|---|---|---|---|---|---|---|---|
| | | Training Data | | | Test Data | | |
| | | Normal | Pathologic | Suspect | Normal | Pathologic | Suspect |
| SVM | Normal | 1131 | 0 | 2 | 520 | 0 | 2 |
| | Pathologic | 0 | 128 | 0 | 0 | 46 | 2 |
| | Suspect | 1 | 0 | 201 | 4 | 2 | 87 |

**Table 4.** Values of statistical measures of the SVM model for training and test data partition before feature selection.

| Model | Partition | Accuracy (%) |
|---|---|---|
| SVM (C = 3, degree = 10) | Training | 99.79 |
| | Test | 98.49 |

Each cell in **Table 5** contains the number of correctly classified instances and incorrectly classified instances corresponding to the percentage of classification accuracy. **Table 6** shows the average classification accuracy achieved by the SVM classifier with polynomial kernel function for C = 10, 50, 100, 500, 1000 and 10,000 with degrees-2, 3, 4, 5 and 10 respectively.

**Table 5** shows the classification accuracy obtained by using a single SVM classifier with the value of C = 10 and degree-3 on the original features. **Table 6** summarizes the classification performance in terms of accuracy for the two parameters tuning (C and degree) of the SVM classifier. From the results we can see that the SVM classifier for the value of C = 10 and degree-3 polynomial has the highest classification accuracy (99.39%) over other values of C and degree parameters. This suggests that a degree-3 polynomial with C = 10 of SVM classifier is flexible enough to discriminate decision boundary. The classification accuracy of SVM with degree-4 polynomial performed worst while the accuracy of degree-5 polynomial yielded a similar classification accuracy of degree-10 polynomial. This shows that increasing the feature space, and in the process increasing the information available for the classifier, does not necessarily improve the accuracy.

### 3.2. Performance Evaluation of Feature Selection with SVM on CTG Data

Our second set of experiments was focused on evaluating the performance of various feature selection techniques with the Support Vector Machine algorithm. The four feature selection techniques *i.e.* Correlation-based feature selection (CorrFS), Consistency-based feature selection (ConsFS), *ReliefF*, and Information Gain (IG) have been applied. We applied best first search strategy, genetic search strategy and ranking method to feature selection. **Table 7** shows the average classification accuracy achieved by the SVM algorithm with polynomial kernel function for C = 10, 100, 500, 10000 and degrees-3, 4, and 5. The table shows sets of experiments each corresponding to a different value of C and degree of SVM classifier.

**Table 5.** Confusion matrix of the SVMs classifier model before feature selection.

| | | | Accuracy (%) |
|---|---|---|---|
| SVM Model (C = 3, degree = 10) | Correctly Classified Instances | 2,113 | 99.39 |
| | Incorrectly Classified Instances | 13 | 0.61 |

**Table 6.** Results of the SVM classifier with polynomial kernel function (C-10, degree-3).

| **Support Vector Machine Classifier Model** | | | | | | | | | |
|---|---|---|---|---|---|---|---|---|---|
| **Degree = 2** | | **Degree = 3** | | **Degree = 4** | | **Degree = 5** | | **Degree = 10** | |
| *C* | Accuracy (%) | *C* | Accuracy (%) | *C* | Accuracy (%) | *C* | Accuracy (%) | *C* | Accuracy (%) |
| 10 | 98.78 | 10 | **99.39** | 10 | 99.20 | 10 | 99.06 | 10 | 99.06 |
| 50 | 99.25 | 50 | 99.29 | 50 | 98.97 | 50 | 99.06 | 50 | 99.06 |
| 100 | 99.34 | 100 | 99.20 | 100 | 98.97 | 100 | 99.06 | 100 | 99.06 |
| 500 | 99.25 | 500 | 99.06 | 500 | 98.97 | 500 | 99.06 | 500 | 99.06 |
| 1000 | 99.15 | 1000 | 99.11 | 1000 | 98.97 | 1000 | 99.06 | 1000 | 99.06 |
| 10,000 | 99.20 | 10,000 | 99.11 | 10,000 | 98.97 | 10,000 | 99.06 | 10,000 | 99.06 |

**Table 7.** Comparison of SVM before feature selection on the CTG data.

| **SVM (Polynomial Kernel)** | | **Accuracy (%)** | | | | | | |
|---|---|---|---|---|---|---|---|---|
| ***C*** | ***degree*** | **Before Feature Selection** | **CorrFS-Best First Search** | **CorrFS-Genetic Search** | **ConsFS-BestFirst Search** | **ConsFS-Genetic Search** | ***ReliefF*-Ranker** | **IG-Ranker** |
| 10 | 3 | **99.39** | 96.14 | 97.93 | 97.18 | 97.55 | **99.39** | **99.39** |
| 500 | 3 | 99.06 | 97.74 | 98.73 | **98.64** | **99.11** | 99.06 | 99.06 |
| 500 | 4 | 98.97 | 98.12 | **99.01** | 98.49 | 98.97 | 98.97 | 98.97 |
| 100 | 5 | 99.06 | 98.02 | **99.01** | 98.21 | 98.73 | 99.06 | 99.06 |
| 10,000 | 5 | 99.06 | **98.4** | 99.06 | 97.55 | 98.82 | 99.06 | 99.06 |

By looking at these results across the different feature selection schemes, two key observations may be made. First, the SVM classifier with feature selection used feature ranking technique *i.e.* Relief and Information Gain outperforms the other two schemes on feature subset selection technique *i.e.* Correlation-based feature selection and Consistency-based feature selection for the value of C = 10 and degree-3 polynomial with 99.39% accuracy. This suggests that the SVM learning based on feature ranking technique is better than the SVM learning based on feature subset selection technique. Second, the correlation-based feature selection and correlation-based feature selection using genetic strategy technique tend to perform more accurate classifiers than the correlation-based feature selection using best first search strategy technique. This suggests that the genetic strategy on feature selection can lead to improved performance.

### 3.3. Performance Evaluation of Ensemble Feature Selection with SVM on CTG Data

This experiment was focused on evaluating the performance of different ensemble feature selections with a single SVM classifier. We used four feature selection techniques of correlation-based feature selection, consistency-based feature selection, *ReliefF*, and Information Gain. To make it easier to compare the various feature selection methods and ensemble feature selections, we assigned "FS1" corresponding to correlation-based feature selection, "FS2" corresponding to consistency-based feature selection, "FS3" corresponding to *ReliefF*, and "FS4" corresponding to Information Gain. For ensemble purposes, the results shown in the column labeled "EFS12" determine the ensemble of correlation-based feature selection and consistency-based feature selection, for instance. Also the column labeled "EFS412" corresponds to the ensemble of three feature selection techniques: Information Gain, correlation-based feature selection, and consistency-based feature selection. This was done to make it easier to represent the various ensemble feature selection techniques based on a single SVM classifier.

The results shown in **Table 8** correspond to the accuracy achieved by the SVM classifiers based on different ensemble feature selection methods. This table contains various models where the SVM classifier achieved higher accuracy than 90%. For example, the result labeled "EFS41-SVM" determines the accuracy achieved by the SVM algorithm when ensemble learning of feature selection is represented as a combination of Information Gain and correlation-based feature selections.

Two key observations can be made by looking at the results in **Tables 8**. First, the SVM classifier (C=10, degree-3) on original features achieves the highest accuracy amongst all experiments of ensemble feature selection with the SVM classifier on varying values of C and degree. Second, comparing the performance of different ensemble feature selection methods based on the SVM classifier, the results demonstrated that the model of ensemble Information Gain and correlation-based feature selection with SVM classifier is able to obtain higher classification accuracies against the performance achieved by other ensemble feature selection methods. These results suggest that feature selection ensembles based on a single SVM are not always better than a single SVM for every case.

## 3.4. Performance Evaluation of Ensemble Feature Selection with SVM Ensembles on CTG data

The last set of results evaluates the performance achieved by ensemble Information Gain and Correlation-based feature selections with ensemble SVM classifiers via majority voting. **Table 9** shows the classification accuracy by the number of SVM classifiers in the corresponding feature spaces. Note that all the SVM results were obtained using polynomial kernel functions with the value of C = 1000 and degree-4 polynomial. **Figure 6** shows the performance of an ensemble of Information Gain and Correlation-based feature selections based on ensemble SVM classifiers in terms of time consumed and accuracy.

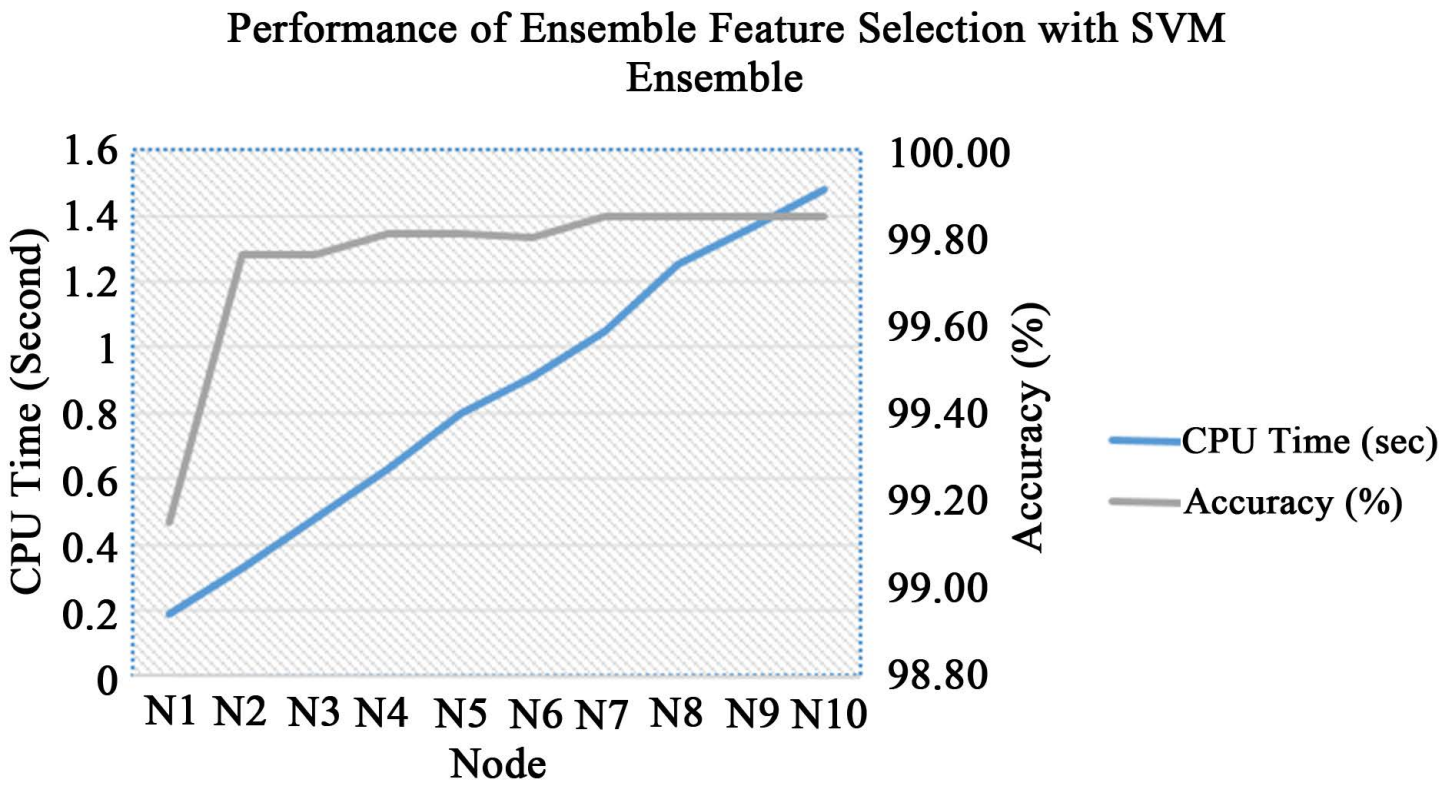

**Figure 6.** Performance of an ensemble of feature selections with SVM ensembles.

**Table 8.** Results of different ensemble feature selection methods with SVM classifiers in terms of accuracy greater than 99%.

| *C* | Degree | Model | Accuracy (%) |
|---|---|---|---|
| 10 | 3 | Original features with SVM classifier | 99.39 |
| 1000 | 4 | EFS41-SVM | 99.15 |
| 500 | 3 | EFS23-SVM, EFS24-SVM | 99.11 |
| 500 | 4 | EFS12-SVM, EFS123-SVM, EFS24-SVM, EFS1234-SVM, EFS13-SVM, EFS14-SVM | 99.01 |
| 100 | 5 | EFS12-SVM, EFS123-SVM, EFS24-SVM, EFS1234-SVM, EFS13-SVM, EFS14-SVM | 99.01 |

**Table 9.** Results of an ensemble of Information Gain feature selection and Correlation-based feature selections with ensemble SVM classifiers (C = 1000, degree-4) for CTG dataset.

| Mem bers | CPU time (sec) | $N_1$ (%) | $N_2$ (%) | $N_3$ (%) | $N_4$ (%) | $N_5$ (%) | $N_6$ (%) | $N_7$ (%) | $N_8$ (%) | $N_9$ (%) | $N_{10}$ (%) | Voting (%) | | |
|---|---|---|---|---|---|---|---|---|---|---|---|---|---|---|
| | | | | | | | | | | | | XS | Agreement | Agreement VS NPS |
| 1 | 0.19 | 99.15 | - | - | - | - | - | - | - | - | - | | | |
| 2 | 0.33 | 99.15 | 98.12 | - | - | - | - | - | - | - | - | 98.92 | 97.79 | 99.76 |
| 3 | 0.48 | 99.15 | 98.12 | 99.06 | - | - | - | - | - | - | - | 99.44 | 97.41 | 99.76 |
| 4 | 0.63 | 99.15 | 98.12 | 99.06 | 98.97 | - | - | - | - | - | - | 99.58 | 96.85 | 99.81 |
| 5 | 0.80 | 99.15 | 98.12 | 99.06 | 98.97 | 98.82 | - | - | - | - | - | 99.48 | 96.52 | 99.81 |
| 6 | 0.91 | 99.15 | 98.12 | 99.06 | 98.97 | 98.82 | 98.82 | - | - | - | - | 99.48 | 96.38 | 99.80 |
| 7 | 1.05 | 99.15 | 98.12 | 99.06 | 98.97 | 98.82 | 98.82 | 98.97 | - | - | - | 99.48 | 95.91 | **99.85** |
| 8 | 1.25 | 99.15 | 98.12 | 99.06 | 98.97 | 98.82 | 98.82 | 98.97 | 99.2 | - | - | 99.44 | 95.86 | **99.85** |
| 9 | 1.36 | 99.15 | 98.12 | 99.06 | 98.97 | 98.82 | 98.82 | 98.97 | 99.2 | 99.2 | - | 99.58 | 95.77 | **99.85** |
| 10 | 1.48 | 99.15 | 98.12 | 99.06 | 98.97 | 98.82 | 98.82 | 98.97 | 99.2 | 99.2 | 99.06 | 99.53 | 95.63 | **99.85** |

**Table 9** records the running time and the number of members for ensemble SVM classifiers and shows the learning accuracy of ensemble SVM on an ensemble of feature selections of Information Gain and Correlation-based. We observe that the running time taken to build the classification model of an ensemble of Information Gain and Correlation-based feature selections is slightly better than the single SVM classifier, in terms of numbers, 1.05 seconds and in terms of learning accuracy on selected feature subset, 99.85%. These results suggest that the ensemble of feature selections with SVM ensembles improve classification accuracy over single SVM classifiers.

## 3.5. Comparison Experimental Results on CRG Data Classification

This section presented the comparing of classification accuracy results and time consumed of five sets of experiments on CTG data. The accuracy rate and processing time reduction in classification were used for evaluating classification efficiency. The result of the experiments is shown in **Table 10**. The column labeled "Member" corresponds to the number of nodes when operating the SVM ensembles. The result labeled "SVM" corresponds to the SVM classifier on original features. For the second experiment, the model labeled "FS1-SVM" corresponds to the result of using correlation-based feature selection with a single SVM classifier; "FS2-SVM" corresponds to the result of using consistency-based feature selection with a single SVM classifier; "FS3-SVM" corresponds to *ReliefF* feature selection with a single SVM classifier; and "FS4-SVM" corresponds to Information Gain feature selection with a single SVM classifier. The model in the third experiment labeled "EFS41-SVM" corresponds to the ensemble of Information Gain feature selection and correlation-based feature selection with a single SVM classifier. The last model labeled "EFS41-ESVM" corresponds to using an ensemble of Information Gain feature selection and correlation-based feature selection with SVM ensembles.

## 3.6. Discussions of the Results and Proposed Model of CTG Data Classification

According to the study on the CTG dataset, Ensemble Information Gain and Correlation-based feature selection with an SVM classifiers ensemble performs the best with regard to accuracy (99.85%). In studying, the results across the different classification schemes we can make two key observations.

First, the performance of the SVM classifier is influenced by the nature of the feature space. This is apparent by comparing the results achieved on the Information Gain's feature space and *ReliefF*'s feature space with the results achieved on the correlation-based's and the consistency-based's feature spaces. The maximum accuracy attained by the SVM classifier on Information Gain and *ReliefF* is always higher than that obtained by SVM on the correlation-based's and the consistency-based's feature spaces. This shows that reducing the size of the feature sets does not necessarily improve the accuracy. At the same time, we see the predictive accuracies achieved from sample sizes without feature selection. These results suggest that feature selection does improve the classification accuracies, but it depends on the method adopted.

Second, we extended the SVM ensembles to the ensemble of Information Gain and Correlation-based feature selection. We observed that the SVM ensembles with ensemble of Information Gain and Correlation-based's feature spaces outperformed a single SVM in terms of classification accuracy. As the size of the ensemble involves balancing speed and accuracy, larger ensembles take longer to train and to generate predictions. Thus, the model of ensemble feature selectin of Information Gain and Correlation-based combined with SVM ensembles with the values of C = 1000, degree-4 can lead to improved performance and be more time-consuming.

Experimenting with these models, we found that the combination of SVM ensemble and ensemble of Information Gain and Correlation-based feature selection model is superior in terms of achieving accuracy improvement. As a result, we are proposing the combination of the ensemble feature selections of Information Gain and Correlation-based with SVM ensembles (C = 1000, degree-4) as our model (**Figure 7**). **Table 11** gives a summary of the parameters setting in our proposed model.

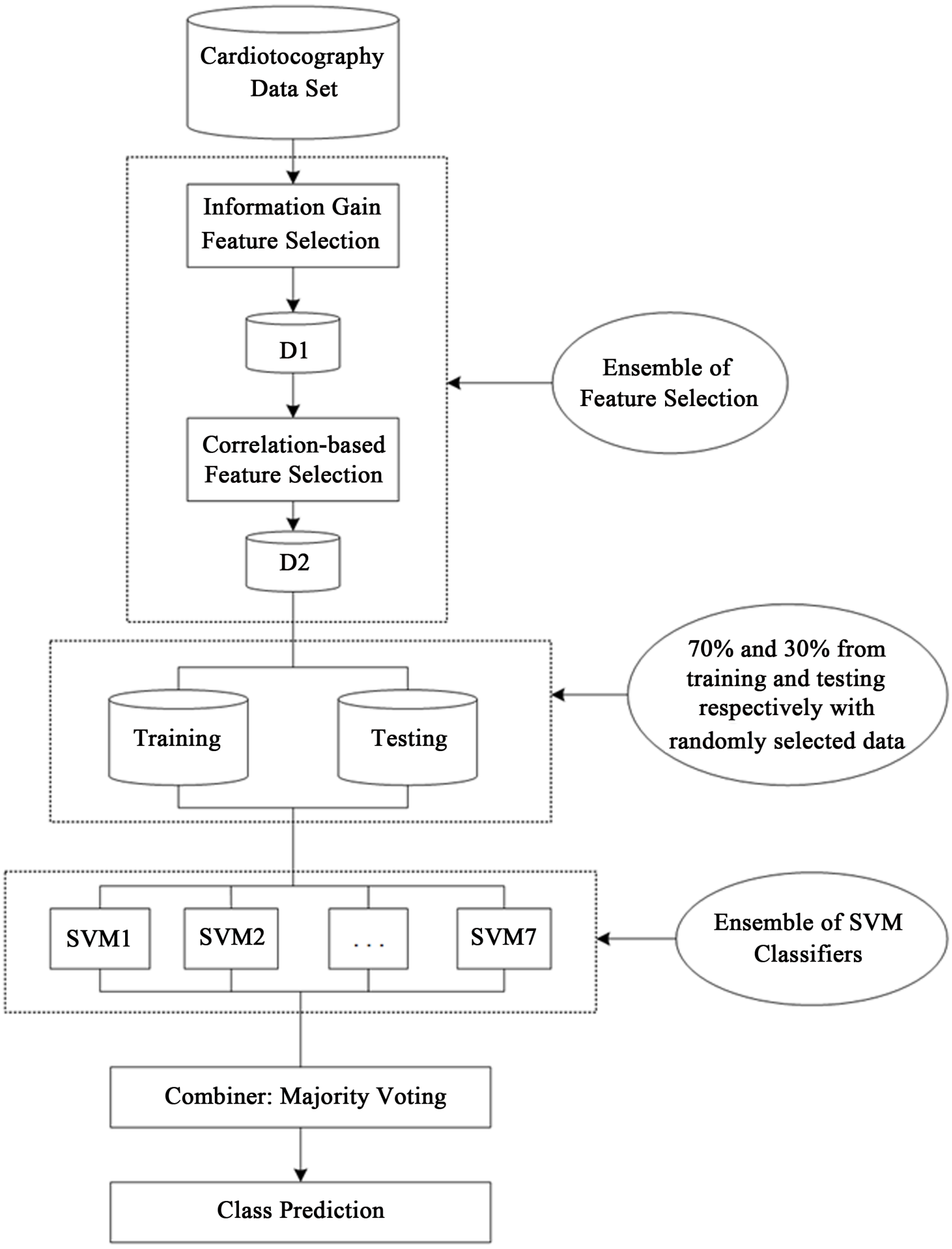

**Figure 7.** Flowchart of the proposed model on CTG data classification.

**Table 10.** Summary of the classification accuracy and CPU time of five sets of experiments on CTG data.

| Experiment | *C* | Degree | Member | Model | Accuracy (%) | CPU Time (s) |
|---|---|---|---|---|---|---|
| 1 | 10 | 3 | - | SVM | 99.39 | 0.42 |
| 2 | 10 | 3 | - | FS1-SVM | 97.93 | 0.20 |
| | 10 | 3 | - | FS2-SVM | 97.55 | 0.27 |
| | 10 | 3 | - | FS3-SVM | 99.39 | 0.44 |
| | 10 | 3 | - | FS4-SVM | 99.39 | 0.42 |
| 3 | 1000 | 4 | - | EFS41-SVM | 99.15 | 0.16 |
| 4 | 1000 | 4 | 7 | EFS41-ESVM | **99.85** | **1.05** |

**Table 11.** Summary of the parameters setting of the proposed model on the CTG data classification.

| Dataset | Feature Selection Methods | SVM (polynomial kernel) | Number of Ensembles SVM |
|---|---|---|---|
| Cardiotocography | Ensemble of Information Gain and Correlation-based | *C*-1000, degree-4 | 7 |

## 4. Conclusions

This paper studies the enhancement of classification accuracy of Cardiotocogram data in ensemble learning based feature selection and classifiers ensemble simultaneously. In feature selection, two filter based feature subset selection techniques (Correlation-based Feature Selection; Consistency-based Filter) and two filter based feature ranking techniques (*ReliefF*; Information Gain) are considered, while Support Vector Machine classification technique is implemented in classification. The performance is evaluated and compared with classification accuracy and the time consumed. We conclude that combining feature selection of Information Gain and Correlation-based with SVM ensembles much improves classification of CTG data.

From experiments of ensemble feature selection with SVM ensembles on the CTG dataset, the first main finding is that the SVM ensemble improves the correct classification rate greatly compared with that of using a single SVM. The second finding is that the majority voting aggregation using the polynomial kernel function with C-1000, degree-4 provides the best classification performance in the CTG dataset. Finally, the classification result of the ensemble of Information Gain and Correlation-based feature selections with the SVM ensemble (C-1000, degree-4) performed with the best accuracy.

The ensemble of Information Gain and Correlation-based feature selections has proved to be suitable for distributing the features among subsets when used with SVM classifier ensemble for combination. The study highlights that the coupling of the method of ensemble feature selection and the classifiers ensemble is crucial for obtaining good results. This proposed approach proves to be able to deal with data challenges, mainly the huge number of features and the small samples size. In addition, the use of the proposed models can be extended to other datasets and other domains in the bioinformatics field.